\documentclass[a4paper]{article}

\usepackage{LaTeX/INTERSPEECH2021}

\usepackage{todonotes}
\usepackage{hyperref}
\usepackage{graphicx}

\usepackage{cite}
\usepackage{siunitx}

\usepackage{xcolor}
\usepackage{pgfplots,pgfplotstable}

\usepackage{multirow}

\newcommand\slimeq{\mkern1.5mu{=}\mkern1.5mu}

\title{Aligned Contrastive Predictive Coding}
\name{Jan Chorowski$^{1,2}$, Grzegorz Ciesielski$^1$, Jarosław Dzikowski$^1$, Adrian {\L}a{\'n}cucki$^3$, Ricard Marxer$^4$, Mateusz Opala$^1$, Piotr Pusz$^1$, Paweł Rychlikowski$^1$ and Michał Stypułkowski$^1$}
\address{
  $^1$University of Wroclaw, Poland\\
  $^2$NavAlgo, France\\
  $^3$NVIDIA\\
  $^4$Université de Toulon, Aix Marseille Univ, CNRS, LIS, France}
\email{jan.chorowski@cs.uni.wroc.pl}

\begin{document}
\maketitle
\begin{abstract}




  We investigate the possibility of forcing a self-supervised model trained using a contrastive predictive loss, to extract slowly varying latent representations. Rather than producing individual predictions for each of the future representations, the model emits a sequence of predictions shorter than the sequence of upcoming representations to which they will be aligned. In this way, the prediction network solves a simpler task of predicting the next symbols, but not their exact timing, while the encoding network is trained to produce piece-wise constant latent codes. We evaluate the model on a speech coding task and  demonstrate that the proposed Aligned Contrastive Predictive Coding (ACPC) leads to higher linear phone prediction accuracy and lower ABX error rates, while being slightly faster to train due to the reduced number of prediction heads.
\end{abstract}





\section{Introduction}

Speech representations learned in an unsupervised way should not be limited to better acoustic features for supervised ASR, but also provide a path towards understanding of utterances at a linguistic level by enabling modeling of audio signals at acoustic, phoneme, or word levels. However, these different levels of speech understanding require reasoning at different sampling rates: acoustic features are computed at equally spaced time intervals (e.g., every 10ms), while language models change state on each phoneme occurrence, or every word occurrence.  To capture this behavior, representation learning methods should provide mechanisms that produce variable-rate data representations, which change synchronously with the contents of the utterance, rather than an external clock.

Supervised ASR models allow variable rate data processing: during recognition an HMM~\cite{Rabiner89-ATO,graves_connectionist_2006,graves_sequence_2012} classifies jointly several acoustic frames into a single phoneme. Similarly the attention mechanism~\cite{bahdanau_neural_2014,chorowski_attentionbased_2015} 
allows a phoneme or a character-synchronous decoder network to access acoustic features sampled uniformly in time. However, training these models requires ground-truth transcriptions. 

\begin{figure}
    \centering\includegraphics[width=.95\columnwidth]{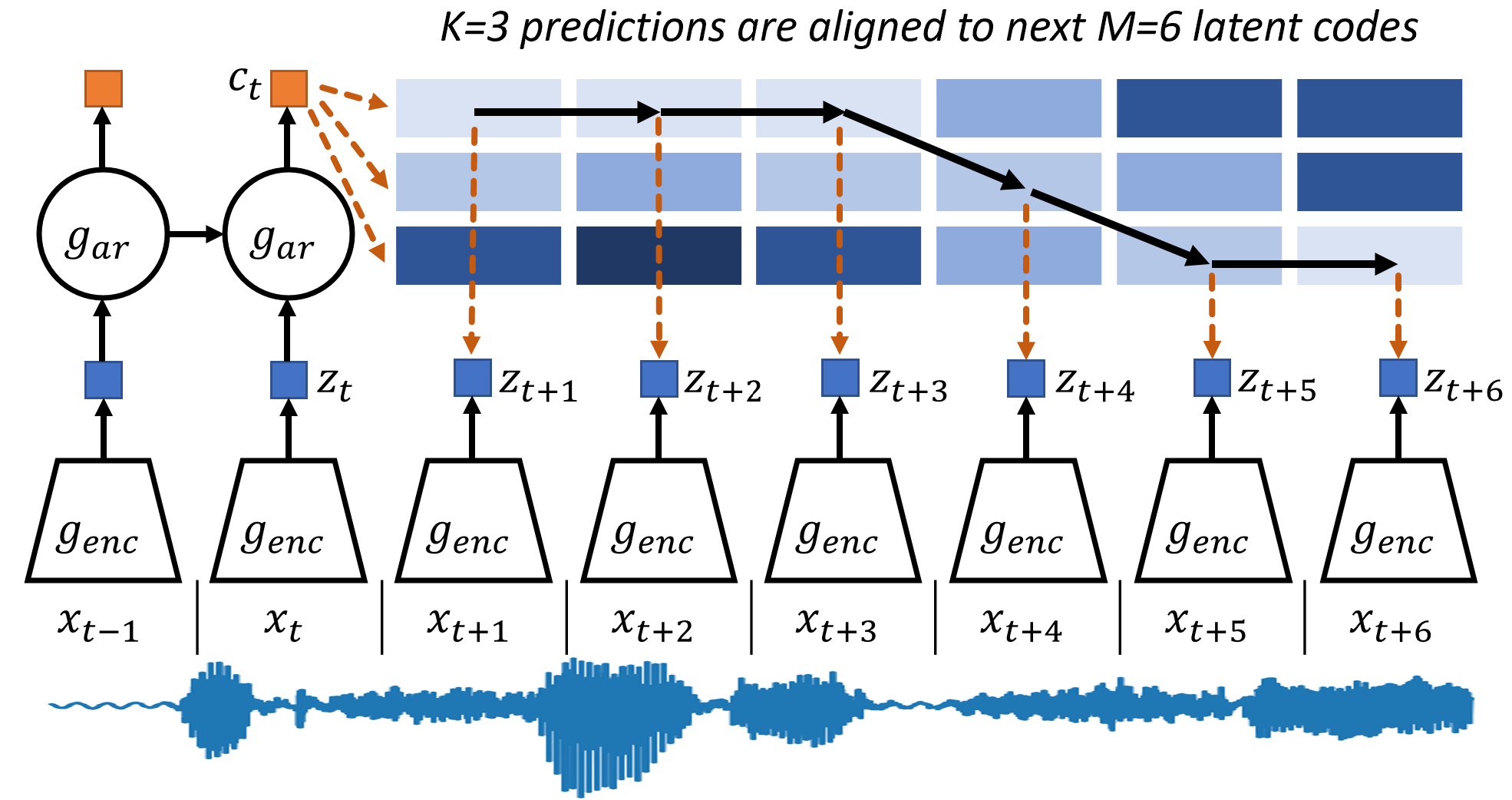}
    \caption{ACPC architecture. The encoder maps chunks of input data into a latent space and the autoregressive model predicts $K$ upcoming latent vectors. They are   aligned using DTW to the $M$ upcoming latent vectors. Training relies on a contrastive loss: the match between a prediction and its aligned latent vectors has to be stronger than the match of the predictor to any other latent vector. Using $K\slimeq M$ is equivalent to CPC~\protect\cite{oord_representation_2018}.}
    \label{fig:acpc}
\end{figure}

In this paper we investigate the possibility of extending  Contrastive Predictive Coding (CPC) \cite{oord_representation_2018}, an unsupervised representation learning method, with a rate-adjusting alignment step. In the proposed Aligned  Contrastive Predictive Coding (ACPC, Figure \ref{fig:acpc}) the model predicts only a short sequence of latent codes which are aligned to a longer series of upcoming ones. 

ACPC changes the behavior of the model in two ways. First, it promotes stability and piece-wise smoothness of the produced encodings: neighboring representations that are aligned to the same prediction are implicitly trained to be similar, while neighboring representations aligned to different predictions are trained to be different. Second, the task learned by the prediction network is altered: rather than predicting precisely timed future representations, the prediction network can now focus on \emph{what} symbols will come next, rather than exactly \emph{when} they will come. In consequence, ACPC leads to representations that are more related to phonemes, both on the linear frame-wise phoneme prediction task and on ABX evaluations. As demonstrated in experiments, ACPC also disincentivises the encoder from extracting location-based features, such as periodical patterns produced by strided convolutions~\cite{odena2016deconvolution} which are induced by classical contrastive coding.

Finally, we note that ACPC brings a small performance increase over CPC. Contrastive scoring of predictions is expensive, because it requires comparisons against hundreds of negative samples. ACPC makes fewer predictions, bringing in a noticeable speedup. 

We publicly release the implementation at \url{https://github.com/chorowski-lab/CPC_audio}.



%


\section{Background and Motivation}

%
%
%

CPC \cite{oord_representation_2018} extracts latent representations of sequential data by learning to predict future states of the model. First, an encoding feed-forward network extracts latent representations from consecutive and overlapping chunks of the data, producing a sequence of latent codes. A powerful auto-regressive model is then applied to the latent representations. It is trained to predict the extracted latent representations several steps into the future. The model is supervised using Noise Contrastive Estimation: the prediction $p_t$ of the latent code $z_t$ at time $t$ must be closer to $z_t$ than to some other randomly sampled latent codes called the negative samples. 
When applied to speech signals, CPC was shown to learn good acoustic representations of the data which are useful for phoneme prediction and low-resource speech recognition.

ACPC makes $K$ predictions of the upcoming latent vectors, which will be then matched to $M$ future latent vectors. It then force-aligns predictions to the latent codes, and only then uses a contrastive loss to supervise the model. The alignment can be performed efficiently using a variant of Dynamic Time Warping (DTW) \cite{vintsyuk_speech_1968}. This extra search step recovers the timing information, freeing the autoregressive model from having to learn it. When $K\slimeq M$ our system is equivalent to the vanilla CPC variant: there is only one possible alignment.

\section{ACPC Details}

Figure~\ref{fig:acpc} 
demonstrates ACPC. The input sequence $x_1,\ldots,x_T$ is mapped using a strided convolutional encoder to a sequence of latent encodings $z_1,\ldots, z_{T'}$. Typically, the encoder uniformly reduces the temporal resolution of the data. Then, the autoregressive model summarizes all encodings up to time $t$, denoted $z_{\leq t}$, into a context vector $c_t = g_{enc}(z_{\leq t})$. Finally, $K$ predictions $p_t^1,\ldots, p_t^K$ are made conditioned on $c_t$. We match these prediction to $M$ upcoming latent vectors $z_{t+1} ,\dots, z_{t+M}$ using contrastive scores as follows. We first sample $N$ negative latent vectors from the encoder outputs $\bar{z}_t^1 ,\dots, \bar{z}_t^N$. We then compute the match between prediction $p_t^k$ and encoding $z_{t+m}$ using the contrastive score:

\begin{equation}\label{eq:scores}
    s_t^{k,m} = \frac{e^{\left<p_t^k, z_{t+m}\right>}}{e^{\left<p_t^k, z_{t+m}\right>} + \sum_{n=1}^N e^{\left<p_t^k, \bar{z}_t^N\right>}},
\end{equation}
where $\left<\cdot,\cdot\right>$ denotes the scalar product of two vectors.

We can organize the scores as a matrix showing the affinity of predictions made at time $t$ to subsequent latent codes. An alignment is a path through this matrix which connects the corner $k\slimeq 1,m\slimeq 1$ with the corner $k\slimeq K,m\slimeq M$. An example is shown in Figure~\ref{fig:acpc}. 
We allow a single prediction to be aligned with several consecutive latent encodings, but each encoding may only be aligned with one prediction.

The model is trained to promote the best alignment between the predictions and actual latent encodings. The alignment can be found using DTW. However, since it is being computed several times for each training minibatch, the computation has to be efficient. We have chosen to approximate DTW using an optimized GPU implementation of CTC \cite{graves_connectionist_2006}. CTC assumes that scores are normalized, and computes the expected cost of the best path. Once a model is trained, a single path is typically dominant, and the expected cost over all paths is close to the cost of the best DTW path. Two technical issues remain. First, CTC uses a blank character. We forbid its use by extending the scoring matrix with a row set to a large negative constant value as its log-probability. Second, CTC requires the scores to be normalized. We use the fact, that when one path dominates, we can add or subtract any number to its scores - and we subtract the softmax normalizing score. The details may be found in the source code.

\section{Experiments}


We conduct our experiments on LibriSpeech train-clean-100, following the setup of \cite{riviere_unsupervised_2020} and the small CPC baseline used in ZeroSpeech 2021 \cite{nguyen_zero_2020}. We keep all design choices, and only replace the CPC loss with ACPC. All models read single channel raw waveforms sampled at 16kHz, chunked into sequences of 20480 samples. The encoder applies five 1D convolutions with internal dimension 256 and filter widths $(10, 8, 4, 4, 4)$. All convolutions are followed by channel-wise magnitude normalization and ReLU activations. Convolutions are strided by $(5, 4, 2, 2, 2)$ respectively, which results in a 160-fold rate reduction, yielding a 256-dimensional latent vector extracted every 10ms. The autoregressive context-building model is a two-layer LSTM network~\cite{hochreiter_long_1997} with 256 hidden units. Finally, each prediction head accesses all past contexts through a single Transformer layer~\cite{vaswani_attention_2017} with 8 scaled dot-product attention heads with internal dimension 2048 and dropout~\cite{srivastava_dropout_2014} with $p=0.1$.

Both CPC and ACPC use 128 negatives. We sample them from other utterances in the same minibatch, and we use minibatches containing 64 utterances of a single speaker. However, there is a subtle interplay between the negative selection and the number of GPUs, that affects the results: the negative selection is restricted to utterances processed on a single GPU. The more GPUs are used, the smaller the pool used for negative selection. In order to simulate the setup of~\cite{nguyen_zero_2020}, we group utterances into 8 groups which we use to simulate 8 GPUs and restrict negative selection to utterances in a group.

\subsection{Qualitative Results}
All results in this section are reported for models trained for 200 epochs. The output of the convolutional encoder ($z_t$ in Figure~\ref{fig:acpc}) is used as the \emph{learned representations} over which all the following experiments are conducted, but similar conclusions apply to  autoregressive contexts $c_t$.  For clarity, we only report ACPC results when $K\slimeq 8$ predictions are matched to a window of $M\slimeq 12$ frames. Several other combinations of predictions and window sizes were also tested ($K\slimeq 10$ over $M\slimeq 16$ and $K\slimeq 12$ over $M\slimeq 20$) leading to similar conclusions. An in-depth investigation of the effects on representation of these settings is left as a scope of future work. 

We investigate three aspects of the learned representations on speech signals: the slow varying temporal evolution, how clustered the representation space is, and the agreement with interpretable views of speech such as the phonetic structure.


\begin{figure}[tb!]
    \centering
    \begin{tabular}{cc}
    \includegraphics[width=.44\columnwidth]{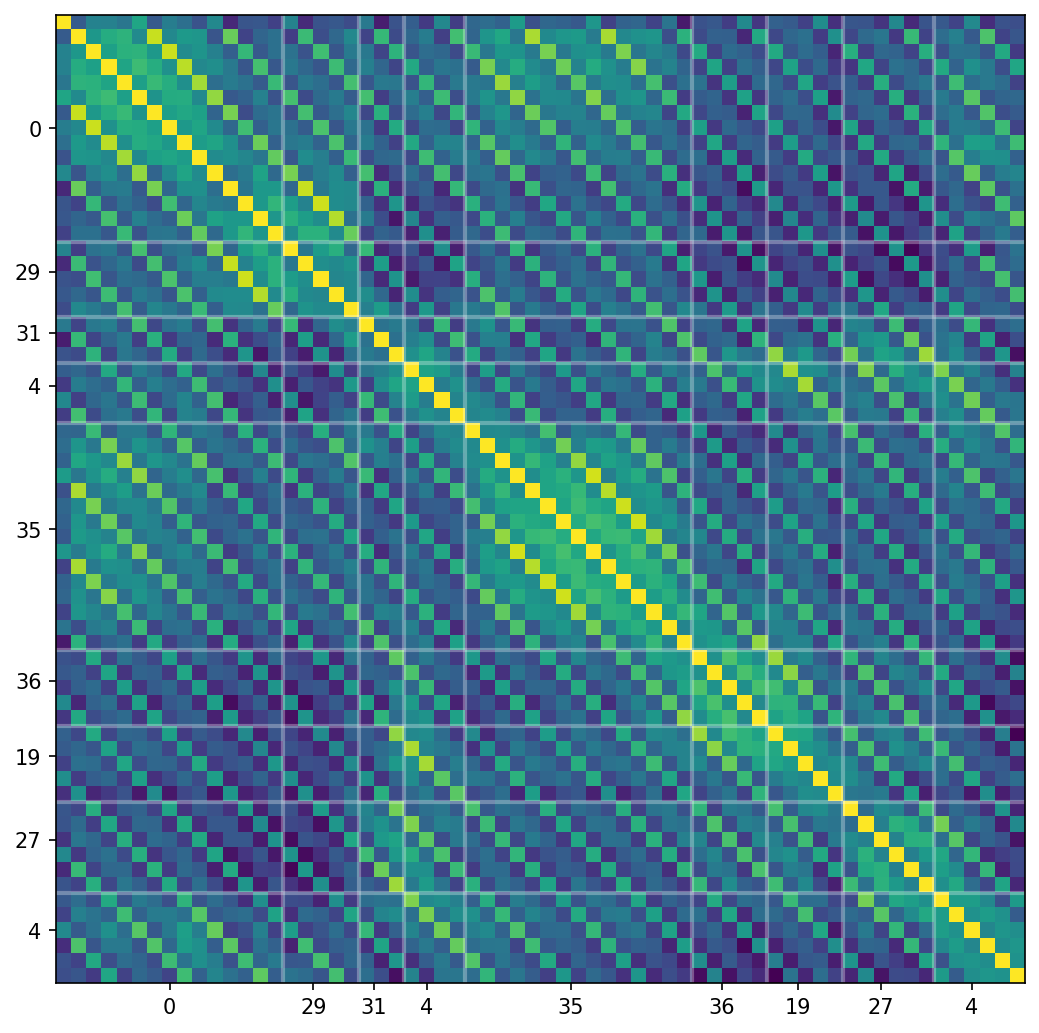} &
    \includegraphics[width=.44\columnwidth]{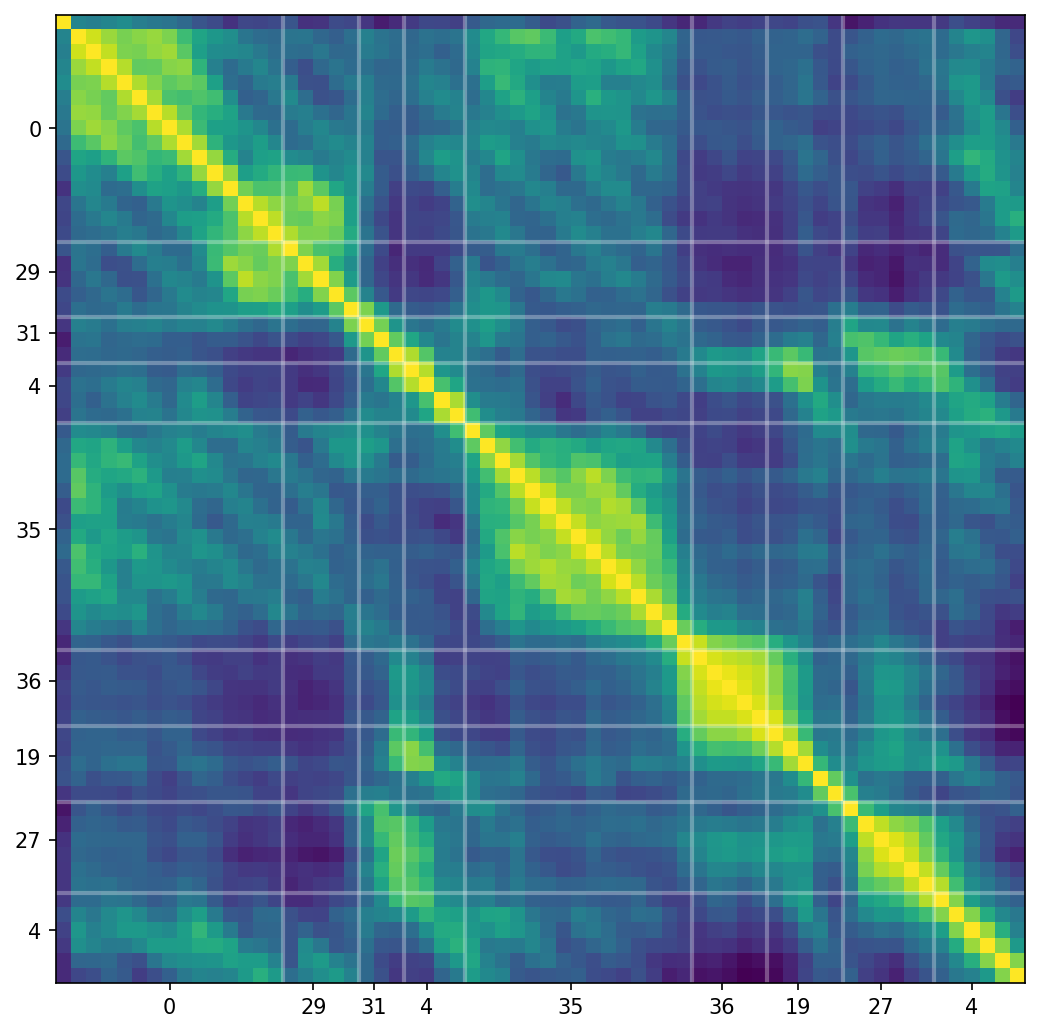}
    \\
    CPC & ACPC (K=8, M=12)
    \end{tabular}%
    \caption{Dot-product similarity of latent representations at the output of the convolutional encoder. The white vertical bars and tick labels indicate ground-truth phoneme segmentation. Moire-like artifacts are visible for CPC, but less pronounced for ACPC (see main text for details). The pronounced blocks on the diagonal align with human-annotated phonemes. Visually, ACPC yields a representation in which latent vectors that correspond to a single phone are more self-similar.}
    \label{fig:enc_simi}
\end{figure}

\begin{figure}[tb]
    \includegraphics[width=\linewidth]{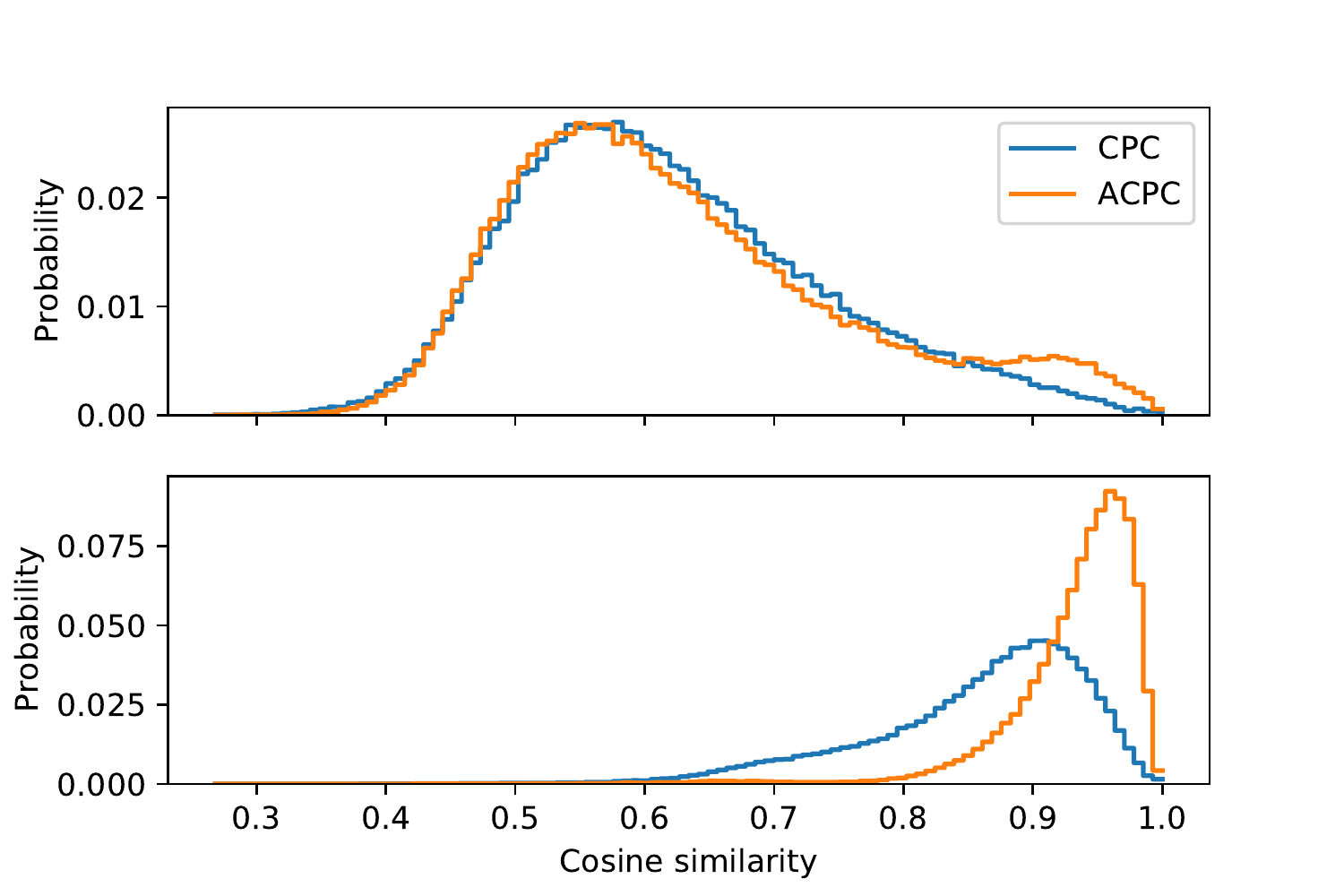}
    \caption{Distribution of cosine similarities between representations of random pairs of frames (top) and consecutive frames (bottom). The histograms are computed on both the representations from the baseline CPC and those of the proposed ACPC.}
    \label{fig:hist_dist}
\end{figure}

\vspace{1ex}\noindent\textbf{Temporal Smoothness } We first compare the self-similarity of representations produced by the convolutional encoder (Figure~\ref{fig:enc_simi}). CPC produces a distinct, moire-like pattern, and we hypothesize that the network learns to abuse strided convolutions to encode the location of the latent vector modulo 4, similarly to the checkerboard pattern visible in deconvolutional image generators~\cite{odena2016deconvolution}. This artifact appears during CPC training because it helps the predictor to guess the upcoming exactly timed features. ACPC does not require exact timing of predictions, and produces only a faint pattern. It enhances the similarity of encodings, as the boundaries of phonemes roughly align with the sharp blocks on the diagonal. We observe this difference between CPC and ACPC repeatedly on other samples.

\vspace{1ex}\noindent\textbf{Quantifying Pairwise Similarities }
We compare pairwise similarities of latent representations between CPC and ACPC with histograms (Figure~\ref{fig:hist_dist}). The distribution between similarities for both models (top) has a similar overall shape. For ACPC there is a second, distinct mode, which is expected for a sharp similarity matrix: a ground truth matrix would be block-diagonal with $O(n)$ ones on the diagonal and $O(n^2)$ zeros elsewhere, motivating two modes unequal in size. We take this idea further and compare self-similarities of only consecutive latent representations (placed directly below/above the diagonal in the distance matrix). There is a significant distribution shift in similarities of consecutive frames between CPC and ACPC.
The alignment of fewer predictions to the sequence of upcoming representations forces the ACPC model to share similar representations for consecutive frames, and therefore creates more slow-varying representation.

\begin{figure}[tb]
    \centering
    \includegraphics[width=\linewidth]{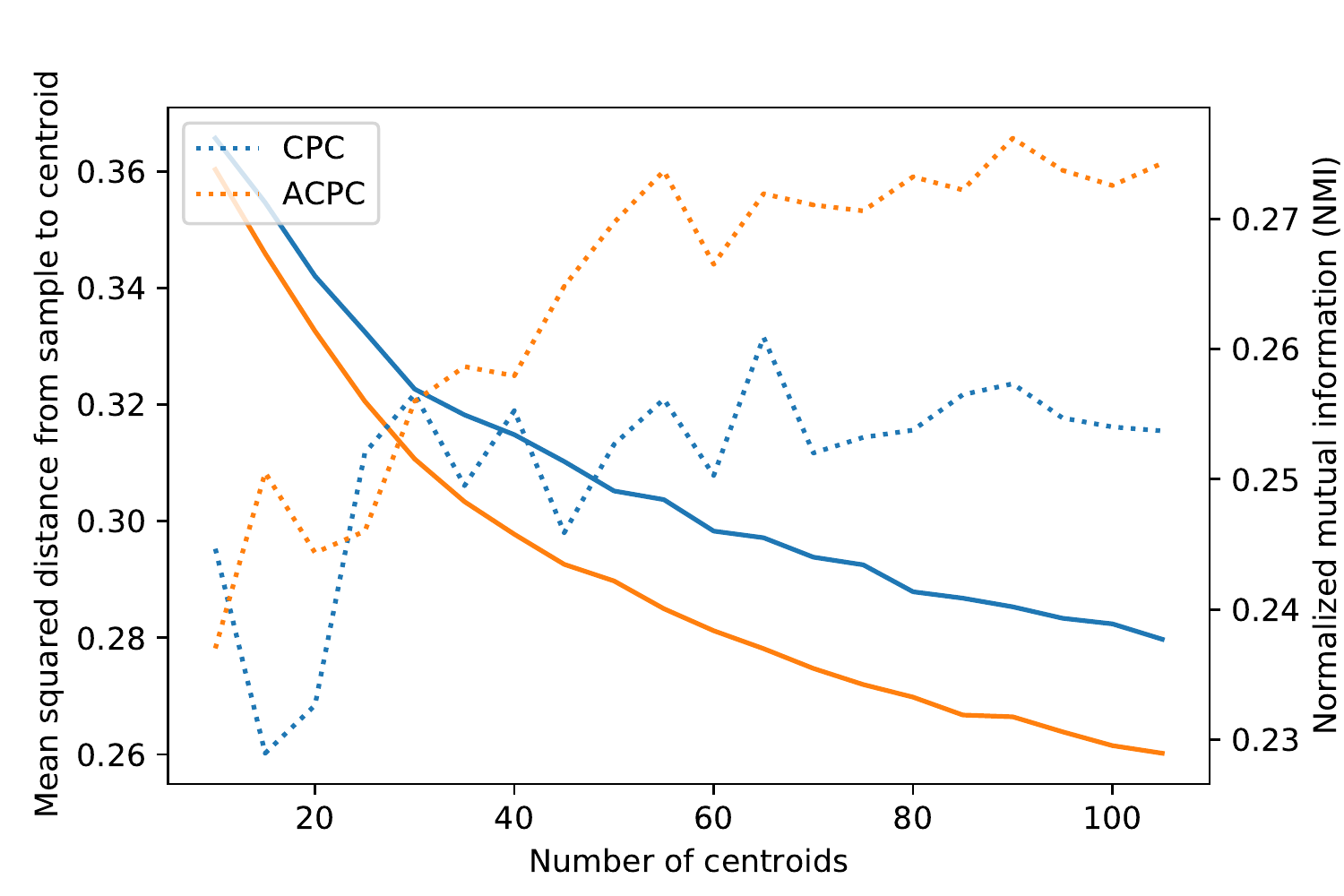}
    \caption{Clustering quality (solid lines, left axis) and agreement with the phone alignments (dashed lines, right axis) of the CPC and ACPC representations for different numbers of centroids. The quality is measured by the average squared distance to the assigned centroid, the agreement  computed by Normalized Mutual Information (NMI) between the cluster assignments and ground truth phones, forced-aligned with ASR. }
    \label{fig:clust}
\end{figure}

\vspace{1ex}\noindent\textbf{Clustering Quality }
While slow-varying features are often expected in speech signals, another aspect which is regularly sought is their readiness to be interpreted categorically. Speech signals are most often segmented into a series of discrete symbols for further processing. Many Automatic Speech Recognition (ASR) perform an alignment of the acoustic stream with regions associated to phone or sub-phone units through classification. We here test how the representations produced by ACPC are clustered in comparison to those of the baseline CPC. We run k-means on the set of latent codes. The clusters are evaluated in terms of concentration by plotting in Figure~\ref{fig:clust} (solid lines and left axis) the average squared distance from the samples to their assigned centroids. We note that the embeddings of both systems have similar norm due to Channel Normalization layers used in the encoder. We observe that ACPC produces consistently a lower mean distance, thus a more clustered representations for every tested number of components.

Furthermore, the clusters created by k-means on the ACPC latents are more in agreement with supervised ASR generated phonetic alignments than those computed using the baseline CPC. We compute and plot
the Normalized Mutual Information 
between the set of cluster assignments resulting from the k-means and the phones assigned to each frame with forced alignment
(Figure~\ref{fig:clust}, dashed lines and right axis).
The results demonstrate that the modes in the latent space of the ACPC map better to phonetic content than those of the baseline CPC.

\subsection{Quantitative Results: ABX and Phone Classification}

We evaluate the model on two tasks that highlight the usefulness of CPC-derived representations: minimal-pair ABX task from ZeroSpeech 2021~\cite{nguyen_zero_2020}, and supervised linear frame-wise phoneme prediction.
For ABX (Table~\ref{tab:abx}), ACPC model is trained with the same settings as the CPC baseline of ZeroSpeech 2021. We extract the features from the last (second) hidden layer of the autoregressive model. We report results for $M\slimeq 12$, $K\slimeq 8$, but similar results are obtained for $K\slimeq 6$. ACPC improves upon the CPC baseline by relative $8-15\%$, depending on the task. It also takes significantly less wall-clock time to train with a similar implementation and hardware.
\begin{table}[tb]
\centering
\caption{ABX error rates on ZeroSpeech 2021 dev set: the provided baseline CPC checkpoint, our CPC rerun and ACPC with M=12 and K=8.}
\label{tab:abx}
\resizebox{\columnwidth}{!}{%
\begin{tabular}{@{\extracolsep{4pt}}lllllll@{}}
  \toprule
  & \multicolumn{2}{l}{CPC (ZeroSpeech \cite{nguyen_zero_2020})} 
  & \multicolumn{2}{l}{CPC (our baseline)}
  & \multicolumn{2}{l}{ACPC M=12 K=8}\\
  \cmidrule{2-3}\cmidrule{4-5}\cmidrule{6-7}
  Dev   & Within & Across & Within & Across & Within & Across \\
  \midrule
  clean & 6.18\% & 8.02\% & 6.68\% & 8.39\% & {\bf 5.37\%} & {\bf 7.09\%} \\
  other & 8.46\% & 13.59\% & 9.03\% & 13.87\% & {\bf 7.46\%} & {\bf 12.60\%} \\
  \bottomrule
\end{tabular}%
}
\end{table}

Next, we provide an ablation on the number of predictions $M$ and predicted frames $K$, with phone classification accuracy task.
We train the models for 50 unsupervised epochs, extract the latent vectors after the convolutional encoder ($z_t$ in Figure~\ref{fig:acpc}) and on the hidden states of the autoregressive model ($c_t$), and then train a supervised linear phone classifier with those representations.
Table~\ref{tab:linphone} lists the results. It consistently indicates that optimal parameters for ACPC in this setup are predicting the next $M\slimeq 12$ frames with $K\slimeq 6$ or $K\slimeq 8$ predictions. We note that the baseline CPC model is set up with $M\slimeq K\slimeq12$.
All tested variants of ACPC improve upon CPC, indicating that $M$ and $K$ are easy to adjust.

\begin{figure}[tb!]
    \centering
    \includegraphics[width=\linewidth]{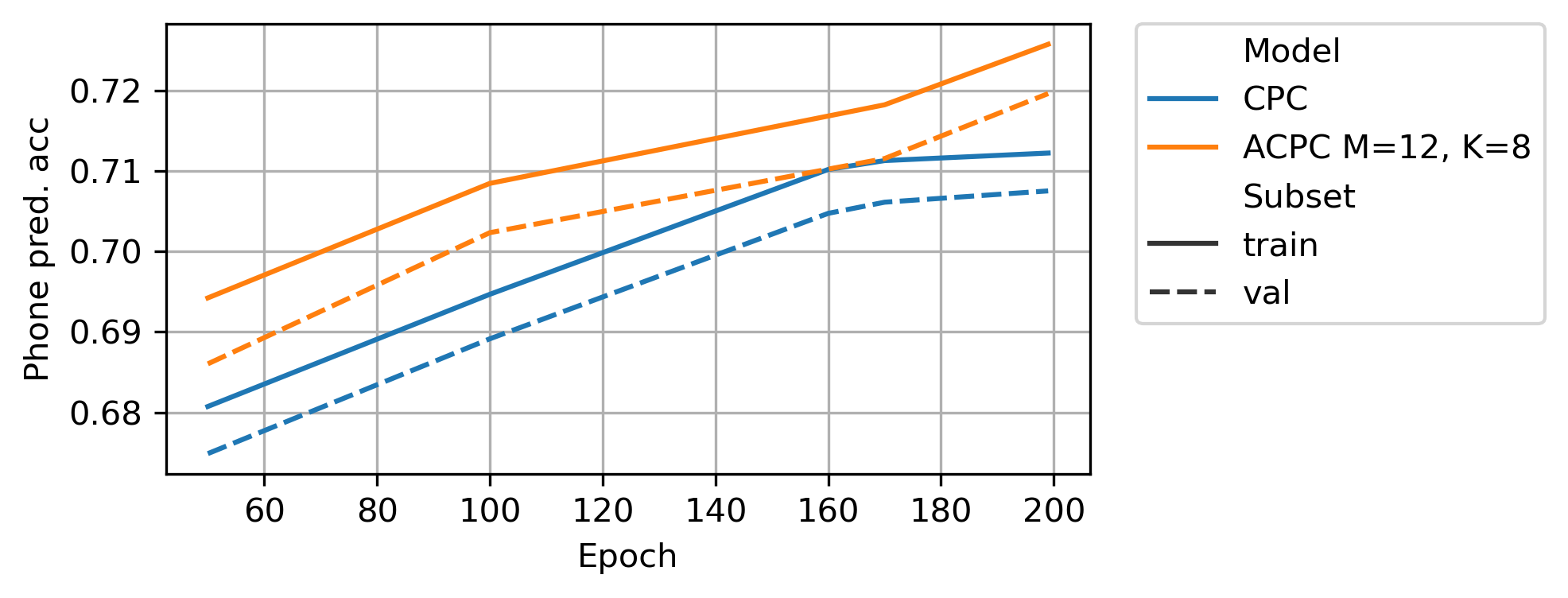}
    \caption{Frame-wise linear phoneme prediction accuracy during training. ACPC steadily improves upon CPC, providing representations
             easier to classify.}
    \label{fig:convergence}
\end{figure}

We measure training step times, collected with the underlying CPC implementation provided for ZeroSpeech 2021. We can see that the training time depends on the amount of predictions, because scoring them is costly - they have to be matched to 128 negative encodings. Thanks to reduced number of predictions, the fastest ACPC setting is 1.7x faster than CPC, and still more accurate on phone classification. Moreover,
ACPC representations are easier to classify from the early training iterations
(Figure~\ref{fig:convergence}).

\begin{table}[tb]
    \centering
    \caption{Frame-wise linear phoneme recognition accuracy after 50 epochs of unsupervised training. We report accuracy obtained on latent representations: outputs $z_t$ of the convolutional encoder, and autoregressive contexts $c_t$ (see Figure~\ref{fig:acpc}). Best results are achieved for M=12 with K=6 or K=8. We report wall-clock training step time.}
    \label{tab:linphone}
    \resizebox{\columnwidth}{!}{%
    \begin{tabular}{@{\extracolsep{4pt}}lcllll@{}}
    \toprule
                   & \multirow{2}{0.72cm}[-0.17cm]{Step time}  & \multicolumn{2}{c}{Encoded $z_t$}  & \multicolumn{2}{c}{Contextual $c_t$}  \\
                   \cmidrule{3-4} \cmidrule{5-6}
    Model          &       & Train   & Val     & Train   & Val  \\
    \midrule
    CPC            & 2.6ms       & 51.5\%       & 51.2\%        & 67.8\%      & 67.5\% \\
    ACPC M=12 K=10 & 2.4ms       & 51.8\%       & 51.3\%        & 68.8\%      & 68.4\% \\
    ACPC M=12 K=8  & 2.1ms       & 52.0\%       & 51.9\%        & {\bf 69.7\%}& 68.6\% \\
    ACPC M=12 K=6  & 1.8ms       & {\bf 52.2\%} & {\bf 52.1\%}  & 69.2\%      & {\bf 68.8\%} \\
    ACPC M=12 K=4  & {\bf 1.5ms} & 52.1\%       & 51.9\%        & 69.0\%      & 68.7\% \\
    ACPC M=16 K=10 & 2.4ms       & 52.1\%       & 51.9\%        & 69.1\%      & {\bf 68.8\%} \\
    ACPC M=20 K=12 & 2.6ms       & 52.1\%       & 51.9\%        & 68.0\%      & 67.8\% \\
    \bottomrule
    \end{tabular}%
    }
\end{table}

We conclude that ACPC is able to train more accurate models. Qualitatively, the learned representations are smoother and more self-similar with frames belonging to each phone. These characteristics translate to improved linear classification accuracy and lower ABX error rates. 
%
%
%
%
%

\section{Related Work}

Slow-feature analysis \cite{wiskott_slow_2002} introduced the general idea that interesting features of a time sequence should be slowly varying and proposed to train models with a penalty on the rate of representation change applied to all subsequent pairs of representations.
We are inspired by this observation, but attempt instead to learn a representation which is piece-wise constant: ideally, the representation would change abruptly at phoneme boundaries, and be fairly stable within a phoneme. In ACPC this is achieved by aligning the representation to a smaller sequence of predictions. Forcibly, some neighboring encodings will be aligned to the same prediction, and thus trained to be more similar, while other neighboring encodings will be aligned to different predictions, and hence trained to be different.

Several solutions were recently proposed to impose slow changes of the latent representations learned in an unsupervised way. So far, most efforts were aimed at auto-encoding models, especially using discrete encodings such as VQ-VAE~\cite{vandenoord_neural_2017}. In \cite{chorowski_unsupervised_2019} a time-jitter regularization was proposed to smooth the latent layer of VQ-VAE, which resulted in improved performance on ZeroSpeech data. \cite{chorowski_unsupervised_2019a} applied a penalty to VQ-VAE latent codes to enforce a piecewise-constant representation with a bounded number of code changes. To this end an optimal cluster assignment problem was solved for each training sample. This approach was refined in \cite{kamper_unsupervised_2020} for phoneme segmentation in a VQ-VAE and VQ-CPC \cite{niekerk_vectorquantized_2020}. Finally, \cite{dieleman_variablerate_2021} have proposed to use slowness penalty and run-length encoding of the latent representation of a VQ-VAE. ACPC is inspired by these solutions, but differs in a fundamental way. All above approaches essentially rely on temporal differences between subsequent latent encodings to enforce a segmentation. This can be seen as a purely bottom-up approach: the segmentation is fully determined given the states of the encoder. ACPC instead uses a top-down signal coming from the predictive network: it aligns the latent encodings to predictions. This makes ACPC somewhat similar to an online variant of pseudo-labeling \cite{likhomanenko_slimipl_2020}: the prediction network produces pseudolabels which after a force-alignment serve as targets for the encoder. \cite{khurana_convolutional_2020} can be seen as another top-down approach which models Markov-based latent transitions and emissions with neural networks, in contrast ACPC does not model explicitly the representation dynamics.

Matching two unaligned sequences of latent states using time warping was explored in the context of ASR by the neural transducer \cite{graves_sequence_2012}. ACPC employs a similar approach for each prediction window.




\section{Conclusion}
We have proposed a modification to the CPC loss which aligns a small sequence of predictions of the upcoming latent vectors to a longer sequence of encoder outputs, then reinforces this alignment using a contrastive training criterion. ACPC changes the training signal to the encoder: it enforces similarity between latent vectors aligned to the same prediction and the task solved by the prediction network: it avoids requiring exact timings of predictions. These changes lead to more stable latent encodings which yield better phoneme prediction accuracies and lower ABX error rates. We treat ACPC as a step towards enforcing meaningful segmentation, such as detecting phoneme boundaries, in a contrastive coding regime. We hope it will be useful in building systems that detect without supervision entities such as phonemes, which are important for a given problem domain. 

\section{Acknowledgments}
The authors thank Polish National Science Center for funding
under the OPUS-18 2019/35/B/ST6/04379 grant and the PlGrid consortium for computational resources.

\bibliographystyle{IEEEtran}
\bibliography{mybib}

\end{document}